\pdfoutput=1

\documentclass[a4paper, 10pt, conference]{ieeeconf}      

\IEEEoverridecommandlockouts                              
\usepackage{cite}

  \usepackage[pdftex]{graphicx}
  \DeclareGraphicsExtensions{.pdf,.jpeg,.png}

\usepackage[cmex10]{amsmath}

\usepackage{mdwmath}
\usepackage{mdwtab}

\usepackage[caption=false]{subfig}

\usepackage{url}

\usepackage{amssymb}
\usepackage[utf8]{inputenc}
\usepackage{cleveref}
\usepackage{booktabs}
\usepackage{threeparttable}
\usepackage{pifont}

\usepackage[nolist,nohyperlinks]{acronym}
\newacro{CNN}{Convolutional Neural Network}
\newacro{GPS}{Global Positioning System}
\newacro{GNSS}{Global Navigation Satellite System}
\newacro{NLOS}{non-line-of-sight}
\newacro{ADAS}{Advanced Driver Assistance Systems}
\newacro{LIDAR}[LiDAR]{Light Detection And Ranging}
\newacro{HD map}{High Definition map}

\def\etal{\emph{et al.}}
\def\eg{\emph{e.g., }}
\def\ie{\emph{i.e., }}
\def\wrt{\emph{w.r.t.}}

\usepackage{tikz}
\usepackage{textcomp}
\usepackage{lipsum}

\newcommand\copyrighttext{%
  \footnotesize \textcopyright 2019 IEEE. Personal use of this material is permitted. Permission from IEEE must be obtained for all other uses, in any current or future media, including reprinting/republishing this material for advertising or promotional purposes, creating new collective works, for resale or redistribution to servers or lists, or reuse of any copyrighted component of this work in other works.}
\newcommand\copyrightnotice{%
\begin{tikzpicture}[remember picture,overlay]
\node[anchor=south,yshift=10pt,xshift=10pt] at (current page.south) {\fbox{\parbox{\dimexpr\textwidth-\fboxsep-\fboxrule\relax}{\copyrighttext}}};
\end{tikzpicture}%
}

\begin{document}

\title{CMRNet: Camera to LiDAR-Map Registration}

\author{
    \authorblockN{
    D. Cattaneo\authorrefmark{1},
    M. Vaghi\authorrefmark{1}
    A. L. Ballardini\authorrefmark{2},
    S. Fontana\authorrefmark{1},
    D. G. Sorrenti\authorrefmark{1}
    W. Burgard\authorrefmark{3}
    }
    \authorblockA{\authorrefmark{1}Universit\`a degli Studi di Milano - Bicocca, Milano, Italy}
    \authorblockA{\authorrefmark{2}Computer Science Department, Universidad de Alcal\'a, Alcal\'a de Henares, Spain}
    \authorblockA{\authorrefmark{3}Albert-Ludwigs-Universit\"at Freiburg, Freiburg, Germany}
    \thanks{\authorrefmark{2}The work of A. L.Ballardini has been funded by European Union H2020, under GA Marie Sk\l{}odowska-Curie n. 754382 Got Energy.}
}

\maketitle
\copyrightnotice 

\begin{abstract}
In this paper we present CMRNet, a realtime approach based on a \ac{CNN} to localize an RGB image of a scene in a map built from LiDAR data.
Our network is not trained in the working area, \ie CMRNet does not learn the map.
Instead it learns to match an image to the map.
We validate our approach on the KITTI dataset, processing each frame independently without any tracking procedure.
CMRNet achieves 0.27m and 1.07$^\circ$ median localization accuracy on the sequence 00 of the odometry dataset, starting from a rough pose estimate displaced up to 3.5m and 17$^\circ$.
To the best of our knowledge this is the first CNN-based approach that learns to match images from a monocular camera to a given, preexisting 3D LiDAR-map.
\end{abstract}


\section{Introduction}\label{sec:introduction}

Over the past few years, the effectiveness of scene understanding for self-driving cars has substantially increased both for object detection and vehicle navigation \cite{Vaquero_2017_ECMR,Chen_2017_CVPR}.
Even though these improvements allowed for more advanced and sophisticated \ac{ADAS} and maneuvers, the current state of the art is far from the SAE full-automation level, especially in complex scenarios such as urban areas.
Most of these algorithms depend on very accurate localization estimates, which are often hard to obtain using common \acp{GNSS}, mainly for \ac{NLOS} and multipath issues.
Moreover, applications that require navigation in indoor areas, \eg valet parking in underground areas, necessarily require complementary approaches.

Different options have been investigated to solve the localization problem, including approaches based on both vision and \ac{LIDAR}; they share the exploitation of an a-priori knowledge of the environment in the localization process \cite{7759304,6942558,8206067}.
Localization approaches that utilize the same sensor for mapping and localization usually achieve good performances, as the map of the scene is matched to the same kind of data generated by the on-board sensor.
However, their application is hampered by the need for a preliminary mapping of the working area, which represents a relevant issue in terms of effort both for building the maps as well as for their maintenance.

On the one hand, some approaches try to perform the localization exploiting standard cartographic maps, such as OpenStreetMap or other topological maps, leveraging the road graph \cite{ParraAlonso2012} or high-level features such as lane, roundabouts, and intersections \cite{Flade2016, Raaijmakers2015, Ballardini2019}.
On the other hand, companies in the established market of maps and related services, like \eg HERE or TomTom, are nowadays already developing so-called \acp{HD map}, which are built using \ac{LIDAR} sensors \cite{heremaps}. This allows other players in the autonomous cars domain, to focus on the localization task.

\acp{HD map}, which are specifically designed to support self-driving vehicles, provide an accurate position of high-level features such as traffic signs, lane markings, etc. as well as a representation of the environment in terms of point clouds, with a density of points usually reaching 0.1m.
In the following, we denote as \ac{LIDAR}-maps the point clouds generated by processing data from \acp{LIDAR}.

\begin{figure}[t]
  \begin{center}
  \includegraphics[width=.48\textwidth]{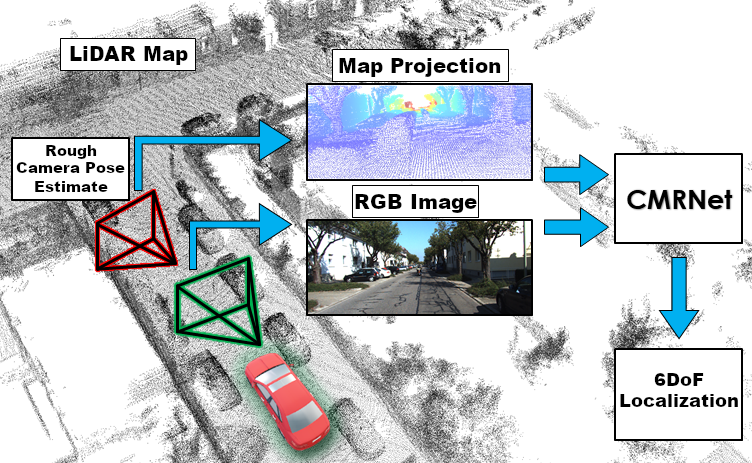}
  \end{center}
  \caption{A sketch of the proposed processing pipeline.
  Starting from a rough camera pose estimate (\eg from a GNSS device), CMRNet compares an RGB image and a synthesized depth image projected from a LiDAR-map into a virtual image plane (red) to regress the 6DoF camera pose (in green).
  Image best viewed in color.
  }
  \label{fig:figure_1_teaser}
\end{figure}

Standard approaches to exploit such maps localize the observer by matching point clouds gathered by the on-board sensor to the \ac{LIDAR}-map; solutions to this problem are known as point clouds registration algorithms.
Currently, these approaches are hampered by the huge cost of \ac{LIDAR} devices, the de-facto standard for accurate geometric reconstruction.


In contrast, we here propose a novel method for registering an image from an on-board monocular RGB camera to a \ac{LIDAR}-map of the area. This allows for the exploitation of the forthcoming market of \ac{LIDAR}-maps embedded into \acp{HD map} using only a cheap camera-based sensor suite on the vehicle.

In particular, we propose CMRNet, a \ac{CNN}-based approach that achieves camera localization with sub-meter accuracy, basing on a rough initial pose estimate.
The maps and images used for localization are not necessarily those used during the training of the network.
To the best of our knowledge, this is the first work to tackle the localization problem without a \emph{localized} \ac{CNN}, \ie a \ac{CNN} trained in the working area \cite{Kendall_2015_ICCV}.
CMRNet does not learn the map, instead, it learns to match images to the \ac{LIDAR}-map.
Extensive experimental evaluations performed on the KITTI datasets \cite{Geiger2013IJRR} show the feasibility of our approach.

The remainder of the paper is organized as follows: \Cref{sec:related-work} gives a short review of the most similar methods and the last achievements with DNN-based approaches.
In \Cref{sec:proposed-approach} we present the details of the proposed system.
In \Cref{sec:experimental} we show the effectiveness of the proposed approach, and \Cref{sec:conclusions,sec:future-works} present our conclusions and future work.

\section{Related work}\label{sec:related-work}
In the last years, visual localization has been a trending topic in the computer vision community.
Although many variations have been proposed, most of them are either based on images gathered from a camera sensor only or exploit some kind of 3-dimensional reconstruction of the environment.

\subsection{Camera-only approaches}
The first category of techniques deals with the 6-DoF estimate of the camera pose using a single image as input.
On the one hand, traditional methods face this problem by means of a two-phase procedure that consists of a coarse localization, performed using a place recognition algorithm, followed by a second refining step that allows for a final accurate localization \cite{Zamir_2010_ECCV, Sattler_2012_ECCV}.
On the other hand, the latest machine learning techniques, mainly based on deep learning approaches, face this task in a single step.
These models are usually trained using a set of images taken from different points of view of the working environment, in which the system performs the localization.
One of the most important approaches of this category, which inspired many subsequent works, is PoseNet \cite{Kendall_2015_ICCV}.
It consists in a \ac{CNN} trained for camera pose regression.
Starting from this work, additional improvements have been proposed by introducing new geometric loss functions \cite{Kendall_2017_CVPR}, by exploiting the uncertainty estimation of Bayesian \acp{CNN} \cite{kendall2016modelling}, by including a data augmentation scheme based on synthetic depth information \cite{Naseer_2017_IROS}, or using the relative pose between two observations in a \acp{CNN} pipeline \cite{Brahmbhatt_2018_CVPR}.
One of the many works that follow the idea presented in PoseNet is VLocNet++ \cite{8458420}.
Here the authors deal with the visual localization problem using a multi-learning task (MLT) approach.
Specifically, they proved that training a \ac{CNN} for different tasks at the same time yields better localization performances than single task learning.
As for today, the literature still sees \cite{8458420} as the best performing approach on the 7Scenes dataset \cite{Shotton_2013_CVPR}.
Clark \etal \cite{Clark_2017_CVPR} developed a CNN that exploits a sequence of images in order to improve the quality of the localization in urban environments.
Brachmann \etal{}, instead, integrated a differentiable version of RANSAC within a CNN-based approach in an end-to-end fashion \cite{Brachmann_2017_CVPR, Brachmann_2018_CVPR}.
Another camera-only localization is based on \textit{decision forests}, which consists of a set of decision trees used for classification or regression problems.
For instance, the approach proposed by Shotton \etal{} \cite{Shotton_2013_CVPR} exploits RGBD images and regression forests to perform indoor camera localization.
The aforementioned techniques, thanks to the generalization capabilities of machine learning approaches, are more robust against challenging scene conditions like lighting variations, occlusions, and repetitive patterns, in comparison with methods based on hand-crafted descriptors, such as SIFT \cite{Lowe2004}, or SURF \cite{BAY2008346}.
However, all these methods cannot perform localization in environments that have not been exploited in the training phase, therefore these regression models need to be retrained for every new place.

\subsection{Camera and \ac{LIDAR}-map approaches}
The second category of localization techniques leverages existing maps, in order to solve the localization problem.
In particular, two classes of approaches have been presented in the literature: geometry-based and projection-based methods.
Caselitz \etal \cite{7759304} proposed a geometry-based method that solves the visual localization problem by comparing a set of 3D points, the point cloud reconstructed from a sequence of images and the existing map.
Wolcott \etal \cite{6942558}, instead, developed a projection-based method that uses meshes built from intensity data associated to the 3D points of the maps, projected into an image plane, to perform a comparison with the camera image using the \textit{Normalized Mutual Information} (NMI) measure.
Neubert \etal \cite{8206067} proposed to use the similarity between depth images generated by synthetic views and the camera image as a score function for a particle filter, in order to localize the camera in indoor scenes.

The main advantage of these techniques is that they can be used in any environment for which a 3D map is available.
In this way, they avoid one of the major drawbacks of machine learning approaches for localization, \ie the necessity to train a new model for every specific environment.
Despite these remarkable properties, their localization capabilities are still not robust enough in the presence of occlusions, lighting variations, and repetitive scene structures.

The work presented in this paper has been inspired by Schneider \etal \cite{Schneider_2017}, which used 3D scans from a \ac{LIDAR} and RGB images as the input of a novel CNN, RegNet.
Their goal was to provide a CNN-based method for calibrating the extrinsic parameters of a camera \wrt~a~\ac{LIDAR} sensor.
Taking inspiration from that work, in this paper we propose a novel approach that has the advantages of both the categories described above.
Differently from the aforementioned literature contribution, which exploits the data gathered from a synchronized single activation of a 3D \ac{LIDAR} and a camera image, the inputs of our approach are a complete 3D \ac{LIDAR} map of the environment, together with a single image and a rough initial guess of the camera pose.
Eventually, the output consists of an accurate 6-DoF camera pose localization.
It is worth to notice that having a single \ac{LIDAR} scan taken at the same time as the image imply that the observed scene is exactly the same.
In our case, instead, the 3D map usually depicts a different configuration, \ie road users are not present, making the matching more challenging.

Our approach combines the generalization capabilities of \acp{CNN}, with the ability to be used in any environment for which a \ac{LIDAR}-map is available, without the need to re-train the network.

\section{Proposed Approach}\label{sec:proposed-approach}
In this work, we aim at localizing a camera from a single image in a 3D \ac{LIDAR}-map of an urban environment.
We exploit recent developments in deep neural networks for both pose regression \cite{Kendall_2015_ICCV} and feature matching \cite{Sun_2018_CVPR}.

The pipeline of our approach is depicted in Fig.~\ref{fig:figure_1_teaser} and can be summarized as follows.
First, we generate a synthesized depth image by projecting the map points into a virtual image plane, positioned at the initial guess of the camera pose.
This is done using the intrinsic parameters of the camera.
From now on, we will refer to this synthesized depth image as \ac{LIDAR}-image.
The \ac{LIDAR}-image, together with the RGB image from the camera, are fed into the proposed CMRNet, which regresses the rigid body transformation $H_{out}$ between the two different points of view.
From a technical perspective, applying $H_{out}$ to the initial pose $H_{init}$ allows us to obtain the 6-DoF camera localization.

In order to represent a rigid body transformation, we use a $(4,4)$ homogeneous matrix:
\begin{equation}
\bf{H} = 
\begin{pmatrix} 
\bf{R}_{(3,3)} & \bf{T}_{(3,1)} \\
\bf{0}_{(1,3)} & \bf{1}
\end{pmatrix}
\in SE(3)
\end{equation}

Here, $\bf{R}$ is a $(3,3)$ rotation matrix and \textbf{T} is a $(3, 1)$ translation vector, in cartesian coordinates.
The rotation matrix is composed of nine elements, but, as it represents a rotation in the space, it only has three degrees of freedom.
For this reason, the output of the network in terms of rotations is expressed using quaternions lying on the 3-sphere ($S^3$) manifold.
On the one hand, even though normalized quaternions have one redundant parameter, they have better properties than Euler angles, \ie gimbal lock avoidance and unique rotational representation (except that conjugate quaternions represent the same rotation).
Moreover, they are composed of fewer elements than a rotation matrix, thus being better suited for machine learning regression approaches.

The outputs of the network are then a translation vector $\textbf{T} \in \mathbb{R}^3$ and a rotation quaternion $\textbf{q} \in S^3$.
For simplicity, we will refer to the output of the network as $H_{out}$, implying that we convert $\textbf{T}$ and $\textbf{q}$ to the corresponding homogeneous transformation matrix, as necessary.

\subsection{LiDAR-Image Generation}\label{sec:lidargeneration}
In order to generate the LiDAR-image for a given initial pose $H_{init}$, we follow a two-step procedure.

\begin{figure}[htb]
\centering
    \subfloat[Without Occlusion Filter]{%
    \label{fig:occlusion_filtering_pre}\includegraphics[width=\linewidth]{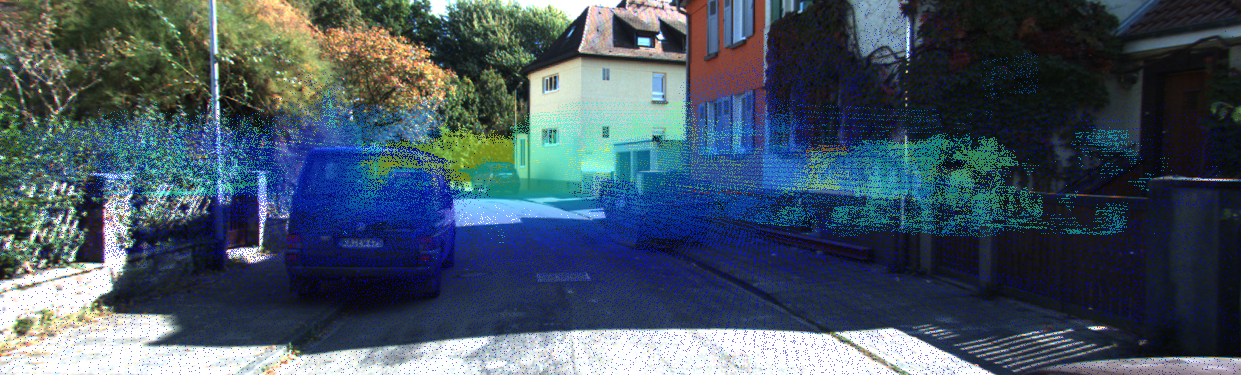}}\hfill
    \subfloat[With Occlusion Filter]{%
    \label{fig:occlusion_filtering_post}\includegraphics[width=\linewidth]{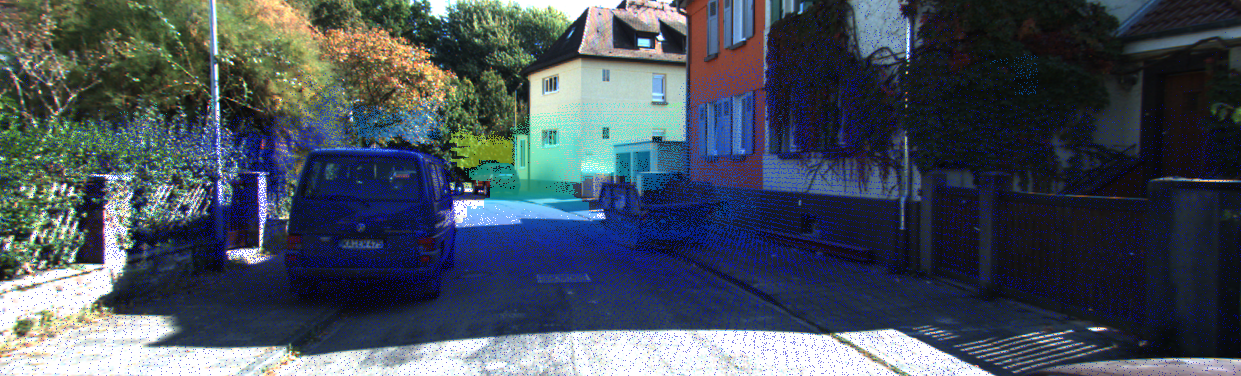}}\hfill
    \caption{Top: a LiDAR-image with the associated RGB overlay.
    Please note how the points behind the building on the right, \ie lighter points on the fence, are projected into the LiDAR-image.
    Bottom: an example of the occlusion filtering effect.
    Color codes distance from close (blue) to far point (red). }\label{fig:occlusion_filtering}
\end{figure}

\noindent
\textit{Map Projection.}
First, we project all the 3D points in the map into a virtual image plane placed at $H_{init}$, \ie compute the image coordinates $p$ of every 3D point $P$.
This mapping is shown in \Cref{eq:proj}, where $K$ is the camera projection matrix.
\begin{equation}
\label{eq:proj}
    p^i = K \cdot H_{init} \cdot P^i
\end{equation}
The LiDAR-image is then computed using a z-buffer approach to determine the visibility of points along the same projection line.
Since \Cref{eq:proj} can be computationally expensive for large maps, we perform the projection only for a sub-region cropped around $H_{init}$, ignoring also points that lay behind the virtual image plane.
In \Cref{fig:occlusion_filtering_pre} is depicted an example of LiDAR-image.


\noindent
\textit{Occlusion Filtering.}
The projection of a point cloud into an image plane can produce unrealistic depth images.
For instance, the projection of occluded points, \eg laying behind a wall, is still possible due to the sparsity nature of point clouds.
To avoid this problem, we adopt the point clouds occlusion estimation filter presented in \cite{pintus2011real}; 
an example of the effect of this approach is depicted in \Cref{fig:occlusion_filtering_post}.
For every point $P_i$, we can build a cone, about the projection line towards the camera, that does not intersect any other point.
If the cone has an aperture larger than a certain threshold \textit{Th}, the point $P_i$ is marked as visible.
From a technical perspective, for each pixel with a non-zero depth $p_j$ in the LiDAR-image, we compute the normalized vector $\Vec{v}$ from the relative 3D point $P_j$ to the pin-hole.
Then, for any 3D point $P_i$ whose projection lays in a neighborhood (of size \textit{KxK}) of $p_j$, we compute the vector ${\Vec{c} = \frac{P_i-P_j}{\|P_i-P_j\|}}$ and the angle between the two vectors ${\vartheta = \arccos(\Vec{v} \cdot \Vec{c})}$.
This angle is used to assess the visibility of $P_j$.
Occluded pixels are then set to zero in the LiDAR-image.
More detail is available in \cite{pintus2011real}

\subsection{Network Architecture}
PWC-Net \cite{Sun_2018_CVPR} was used as baseline, and we then made some changes to its architecture.
We chose this network because PWC-Net has been designed to predict the optical flow between a pair of images, \ie to find matches between them.
Starting from a rough camera localization estimate, our insight is to exploit the correlation layer of PWC-Net and its ability to match features from different points of view to regress the correct 6-DoF camera pose.

We applied the following changes to the original architecture.
\begin{itemize}
    \item First, as our inputs are a depth and an RGB image (instead of two RGB images), we decoupled the feature pyramid extractors by removing the weights sharing.
    \item Then, as we aim to perform pose regression, we removed the up-sampling layers, attaching the fully connected layers just after the first cost volume layer.
\end{itemize}
Regarding the regression part, we added one fully connected layer with 512 neurons before the first optical flow estimation layer (conv6\_4 in PWC-Net), followed by two branches for handling rotations and translations.
Each branch is composed of two stacked fully connected layers, the first with 256 while the second with 3 or 4 neurons, for translation and rotation respectively.

Given an input pair composed of a RGB image $\mathcal{I}$ and a LiDAR-image $\mathcal{D}$, we used the following loss function in \Cref{eq:loss}, where  $\mathcal{L}_t(\mathcal{I}, \mathcal{D})$ is the translation loss and $\mathcal{L}_q(\mathcal{I}, \mathcal{D})$ is the rotation loss.
\begin{equation}
    \label{eq:loss}
    \mathcal{L}(\mathcal{I}, \mathcal{D}) = \mathcal{L}_t(\mathcal{I}, \mathcal{D}) + \mathcal{L}_q(\mathcal{I}, \mathcal{D})
\end{equation}
For the translation we used a $\mbox{smooth}_{L1}$ loss \cite{girshick2015fast}.
Regarding the rotation loss, since the Euclidean distance does not provide a significant measure to describe the difference between two orientations, we used the angular distance between quaternions, as defined below:
\begin{eqnarray}\label{eq:lq}
    \mathcal{L}_q(\mathcal{I}, \mathcal{D}) & = & D_a(q*inv(\Tilde{q})) \\
    \label{eq:atan2}
    D_a(m) &=& atan2(\sqrt{b^2_m + c^2_m + d^2_m}, |a_m|)
\end{eqnarray}
Here, $q$ is the ground truth rotation, $\Tilde{q}$ represents the predicted normalized rotation, $inv$ is the inverse operation for quaternions, $\{a_m, b_m, c_m, d_m\}$ are the components of the quaternion $m$ and $*$ is the multiplicative operation of two quaternions.

In order to use Equation~\eqref{eq:atan2} as a loss function, we need to ensure that it is differentiable for every possible output of the network.
Recalling that $atan2(y,x)$ is not differentiable for ${y=0 \land x\leq0}$, and the fact that $m$ is a unit quaternion, we can easily verify that Equation~\eqref{eq:atan2} is differentiable in $S^3$.

\subsection{Iterative refinement}
\label{seq:iterative}
When the initial pose strongly deviates with respect to the camera frame, the map projection produces a LiDAR-image that shares just a few correspondences with the camera image.
In this case, the camera pose prediction task is hard, because the \ac{CNN} lacks the required information to compare the two points of view.
It is therefore quite likely that the predicted camera pose is not accurate enough.
Taking inspiration from \cite{Schneider_2017}, we propose an iterative refinement approach.
In particular, we trained different CNNs by considering descending error ranges for both the translation and rotation components of the initial pose.
Once a LiDAR-image is obtained for a given camera pose, both the camera and the LiDAR-image are processed, starting from the CNN that has been trained with the largest error range.
Then, a new projection of the map points is performed, and the process is repeated using a CNN trained with a reduced error range.
Repeating this operation \textit{n} times is possible to improve the accuracy of the final localization.
The improvement is achieved thanks to the increasing overlap between the scene observed from the camera and the scene projected in the \textit{$n^{th}$} LiDAR-image.

\subsection{Training details}
We implemented CMRNet using the PyTorch library \cite{paszke2017automatic}, and a slightly modified version of the official PWC-Net implementation.
Regarding the activation function, we used a leaky RELU (REctified Linear Unit) with a negative slope of $0.1$ as non-linearity.
Finally, CMRNet was trained from scratch for 300 epochs using the ADAM optimizer with default parameters, a batch size of $24$ and a learning rate of $1e^{-4}$ on a single NVidia GTX 1080ti.

\section{Experimental results}\label{sec:experimental}
This section describes the evaluation procedure we adopted to validate CMRNet, including the used dataset, the assessed system components, the iterative refinements and finally the generalization capabilities.

We wish to emphasize that, in order to assess the performance of CMRNet itself, in all the performed experiments each input was processed independently, \ie without any tracking or temporal integration strategy.

\subsection{Dataset}
We tested the localization accuracy of our method on the KITTI odometry dataset.
Specifically, we used the sequences from 03 to 09 for training (11697 frames) and the sequence 00 for validating (4541 frames).
Note that the validation set is spatially separated from the train set, except for a very small sub-sequence (approx 200 frames), thus it is fair to say that the network is tested in scenes never seen during the training phase.
Since the accuracy of the provided GPS-RTK ground truth is not sufficient for our task (the resulting map is not aligned nearby loop closures), we used a LiDAR-based SLAM system to obtain consistent trajectories.
The resulting poses are used to generate a down-sampled map with a resolution of 0.1m.
This choice is the result of our expectations on the format of HD-maps that will be soon available from map providers \cite{heremaps}.

Since the images from the KITTI dataset have different sizes (varying from 1224x370 to 1242x376), we padded all images to 1280x384, in order to match the \ac{CNN} architecture requirement, \ie width and height multiple of 64.
Note that we first projected the map points into the LiDAR-image and then we padded both RGB and LiDAR-image, in order not to modify the camera projection parameters.

To simulate a noisy initial pose estimate $H_{init}$, we applied, independently for each input, a random translation, and rotation to the ground truth camera pose.
In particular, for each component, we added a uniformly distributed noise in the range of [-2m, +2m] for the translation and [$-10^{\circ}$, $+10^{\circ}$] for the rotation.

Finally, we applied the following data augmentation scheme: first, we randomly changed the image brightness, contrast and saturation (all in the range [0.9, 1.1]).
Then we randomly mirrored the image horizontally, and last we applied a random image rotation in the range [$-5^\circ$, $+5^\circ$] along the optical axis.
The 3D point cloud was transformed accordingly.

Both data augmentation and the selection of $H_{init}$ take place at run-time, leading to different LiDAR-images for the same RGB image across epochs.


\subsection{System Components Evaluation}
We evaluated the performances of CMRNet by assessing the localization accuracy, varying different sub-components of the overall system.
Among them, the most significative are shown in \Cref{tab:risultatigrossi}, and derive from the following operational workflow.

First, we evaluated the best CNN to be used as backbone, comparing the performances of state-of-the-art approaches, namely PWC-Net, ResNet18 and RegNet \cite{Sun_2018_CVPR, he2016deep, Schneider_2017}.
According to the performed experiments, PWC-Net maintained a remarkable superiority with respect to RegNet and ResNet18 and therefore was chosen as a starting point for further evaluation.

Thereafter, we estimated the effects in modifying both inputs, \ie camera images and \ac{LIDAR}-images.
In particular, we added a random image mirroring and experimented different parameter values influencing the effect of the occlusion filtering presented in \Cref{sec:lidargeneration}, \ie size \textit{K} and threshold \textit{Th}.

At last, the effectiveness of the rotation loss proposed in \Cref{eq:atan2} was evaluated with respect to the commonly used $L_1$ loss. The proposed loss function achieved a relative decrease of rotation error of approx. 35\%.

The noise added to the poses in the validation set was kept fixed on all the experiments, allowing for a fair comparison of the performances.

\begin{table}[]
  \scriptsize
  \centering
  \begin{threeparttable}
    \caption{Parameter Estimation}
    \label{tab:risultatigrossi}
    \begin{tabular*}{\linewidth}{ccccc|cc}
      \toprule
      & \multicolumn{2}{c}{Occlusion} & &  &\multicolumn{2}{c}{\textbf{Error}} \\ \cmidrule(lr){2-3} \cmidrule(lr){6-7}
      Backbone & \textit{K} & \textit{Th} & Mirroring  & Rot. Loss & Transl. & Rot. \\ \midrule
      
      Regnet   & -      &   -       & \ding{55}  & $D_a$&   0.64m   &   1.67$^\circ $ \\
      ResNet18 & -      &   -       & \ding{55}  & $D_a$&   0.60m   &   1.59$^\circ $ \\
      PWC-Net  & 11     &   3.9999  & \ding{55}  & $D_a$&   0.52m   &   1.50$^\circ $ \\
      PWC-Net  & 13     &   3.9999  & \ding{55}  & $D_a$&   0.51m   &   1.43$^\circ $ \\
      PWC-Net  & 5      &   3.0     & \ding{55}  & $D_a$&   0.47m   &   1.45$^\circ $ \\
      PWC-Net  & 5      &   3.0     & \ding{51}  & $D_a$&   \textbf{0.46m}   &   $\textbf{1.36}^\circ$ \\
      PWC-Net  & 5      &   3.0     & \ding{51}  & $L_1$&   0.46m   &   2.07$^\circ$ \\ \bottomrule
    \end{tabular*}
    \begin{tablenotes}[para,flushleft]
      \footnotesize      
      \item Median localization accuracy varying different sub-components of the overall system. \textit{K} and \textit{Th} correspond to the occlusion filter parameters as described in \Cref{sec:lidargeneration}.
    \end{tablenotes}
  \end{threeparttable}
\end{table}

\subsection{Iterative Refinement and Overall Assessment}
In order to improve the localization accuracy of our system, we tested the iterative approach explained in \Cref{seq:iterative}.
In particular, we trained three instances of CMRNet varying the maximum error ranges of the initial camera poses.
To assess the robustness of CMRNet, we repeated the localization process for 10 times using different initial noises. The averaged results are shown in \Cref{tab:iterative} together with the correspondent ranges used for training each network.

Moreover, in order to compare the localization performances with the state-of-the-art monocular localization in \ac{LIDAR} maps \cite{7759304}, we calculated mean and standard deviation for both rotation and translation components over 10 runs on the sequence 00 of the KITTI odometry dataset.
Our approach shows comparable values for the translation component ($0.33\pm0.22$m \wrt~$0.30\pm0.11$m), with a lower rotation errors ($1.07\pm0.77^\circ$ \wrt~$1.65\pm0.91^\circ$).
Nevertheless, it is worth to note that our approach still does not take advantage of any pose tracking procedure nor multi-frame analysis.

\begin{figure*}
  \begin{center}
  \includegraphics[width=1.0\textwidth]{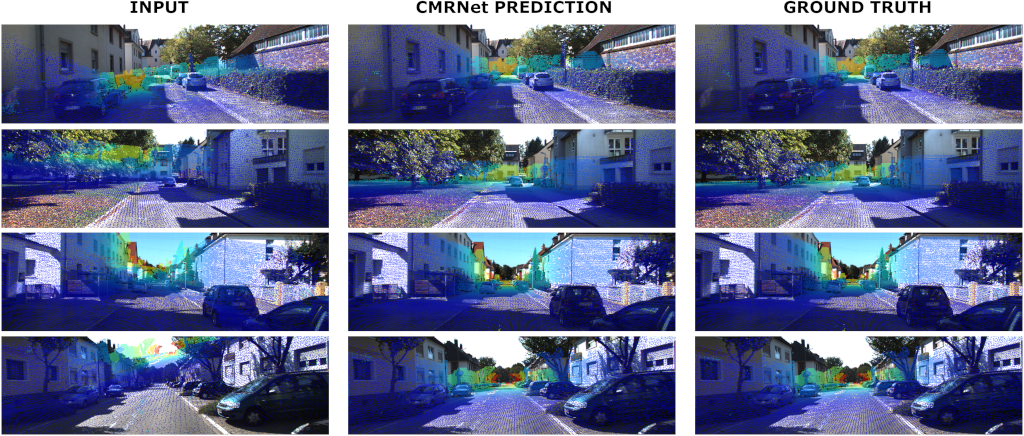}
  \end{center}
  \caption{Four examples of the localization results.
  From left to right: Input LiDAR-image, CMRNet result after the third iteration, ground truth.
  All \ac{LIDAR}-images are overlayed with the respective RGB image for visualization purpose.}
  \label{fig:composition}
  \vspace{-5mm}
\end{figure*}

Some qualitative examples of the localization capabilities of CMRNet with the aforementioned iteration scheme are depicted in \Cref{fig:composition}.

\begin{figure*}[htb]
\centering
    \subfloat[Longitudinal Errors]{%
    \includegraphics[width=.32\textwidth]{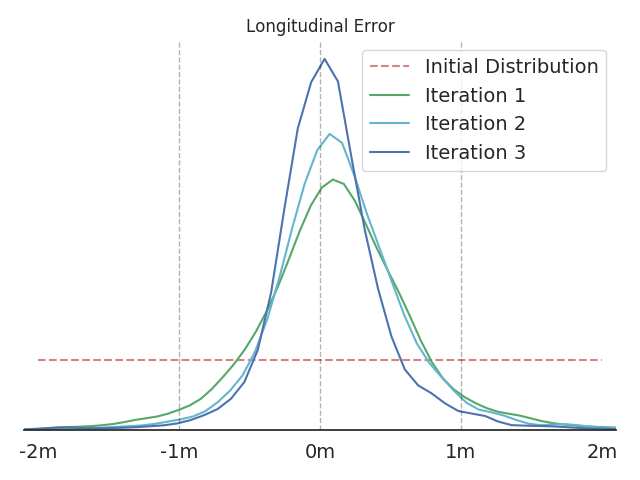}}\hfill
    \subfloat[Lateral Errors]{%
    \includegraphics[width=.32\textwidth]{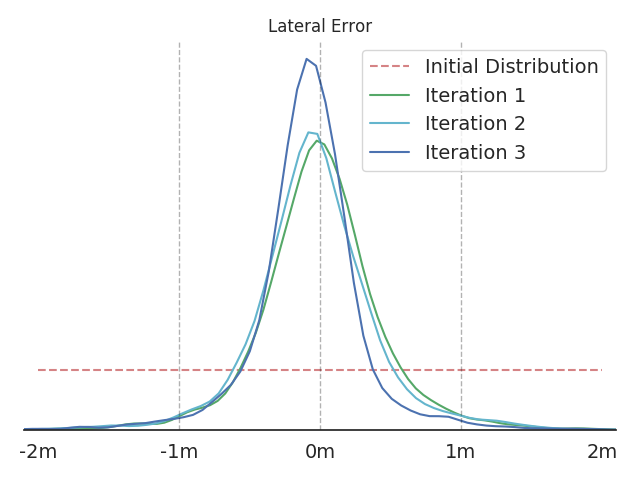}}\hfill
    \subfloat[Vertical Errors]{%
    \includegraphics[width=.32\textwidth]{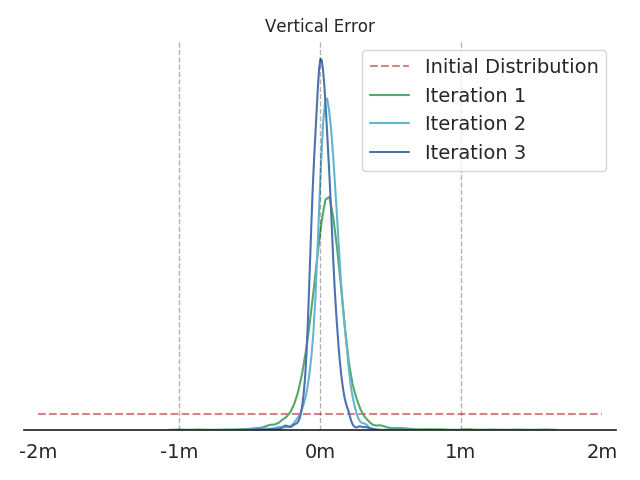}}\hfill
    \subfloat[Roll Errors]{%
    \includegraphics[width=.32\textwidth]{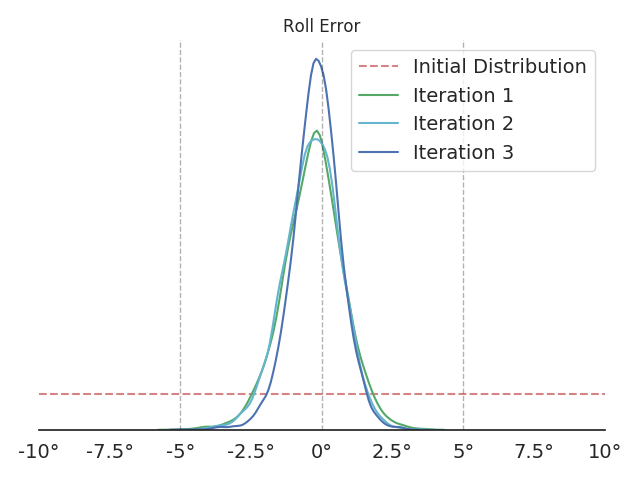}}\hfill
    \subfloat[Pitch Errors]{%
    \includegraphics[width=.32\textwidth]{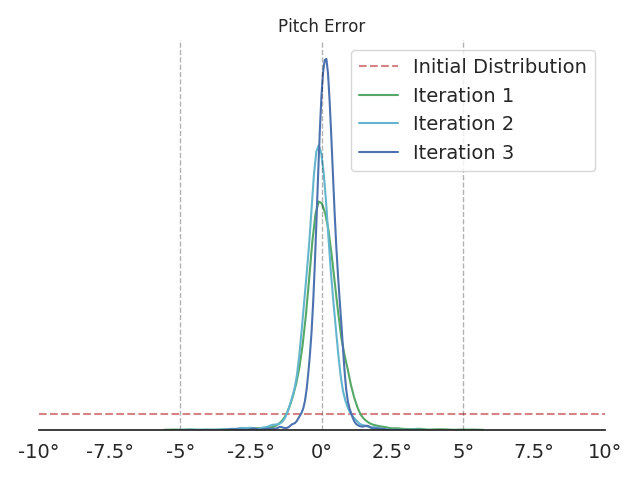}}\hfill
    \subfloat[Yaw Errors]{%
    \includegraphics[width=.32\textwidth]{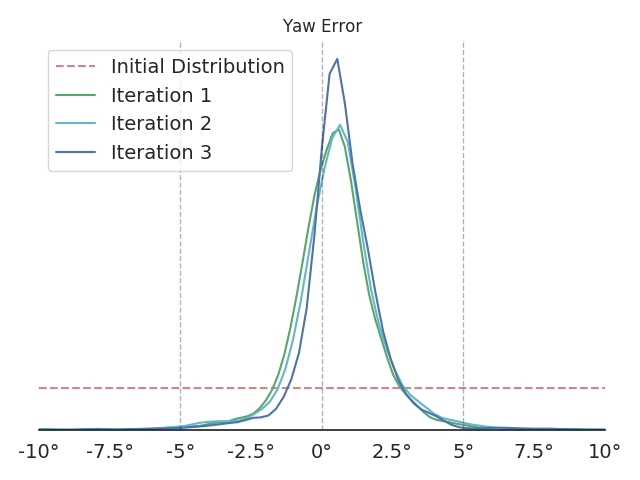}}\hfill
    \caption{Iterative refinement error distributions: a PDF has been fitted (using Gaussian kernel density estimation) on the network error outcome for each iteration step and each component.
    The dashed red lines are the theoretic PDFs of the initial $H_{init}$ errors.}\label{fig:pdf}
    \vspace{-5mm}
\end{figure*}

In \Cref{fig:pdf} we illustrate the probability density functions (PDF) of the error, decomposed into the six components of the pose, for the three iterations of the aforementioned refinement.
It can be noted that the PDF of even the first network iteration approximates a Gaussian distribution and following iterations further decrease the variance of the distributions.

An analysis of the runtime performances using this configuration is shown in \Cref{tab:runtime}.

\begin{table}[]
\centering
\begin{threeparttable}
\caption{Iterative Pose Refinement}
\label{tab:iterative}
\begin{tabular*}{\linewidth}{lcc|cc}
\toprule
 & \multicolumn{2}{c}{Initial Error Range}& \multicolumn{2}{c}{Localization Error} \\ \cmidrule(lr){2-3} \cmidrule(lr){4-5}
            & Transl. [m] & Rot. [deg] & Transl. [m] & Rot. [deg] \\ \midrule
Iteration 1 & [-2, +2]& [$-10$, $+10$] &   0.51         &  $1.39$   \\
Iteration 2 & [-1, +1]& [$-2$, $+2$] &   0.31         &  $1.09$   \\
Iteration 3 & [-0.6, +0.6]& [$-2$, $+2$]&  \textbf{0.27} &  $\textbf{1.07}$   \\ \bottomrule
\end{tabular*}
    \begin{tablenotes}[para,flushleft]
      \footnotesize      
      \item Median localization error at each step of the iterative refinement averaged over 10 runs.
    \end{tablenotes}
\end{threeparttable}
\end{table}

\begin{table}[t!]
\centering
\begin{threeparttable}
\caption{Runtime Performances}
\label{tab:runtime}
\begin{tabular*}{\linewidth}{cccc|c}
\toprule
              & Z-Buffer & Occlusion Filter & CMRNet & \textbf{Total} \\ \midrule
Time {[}ms{]} & 8.6      & 1.4              & 4.6    & \textbf{14.7} ($\sim$68Hz) \\ \bottomrule
\end{tabular*}
    \begin{tablenotes}[para,flushleft]
      \footnotesize     
      \item In the table, an analysis of the time performances of the system steps for a single execution, \ie 44.1ms for the 3-stages iterative refinement.
      All the code was developed in CUDA, achieving 68fps runtime performances on the KITTI dataset.
      CPU-GPU transfer time was not here considerated.
    \end{tablenotes}
\end{threeparttable}
\end{table}


\subsection{Generalization Capabilities}
In order to assess the generalization effectiveness of our approach, we evaluated its localization performance using a 3D \ac{LIDAR}-map generated on a different day with respect to the camera images, yet still of the same environment.
This allows us to have a completely different arrangement of parked cars and therefore to stress the localization capabilities.

Unfortunately, there is only a short overlap between the sequences of the odometry dataset (approx. 200 frames), consisting of a small stretch of roads in common between sequences "00" and "07".
Even though we cannot completely rely on the results of this limited set of frames, CMRNet achieved 0.57m and 0.9$^\circ$ median localization accuracy on this test.

Indeed, it is worth to notice that the network was trained with maps representing the same exact scene of the respective images, \ie with cars parked in the same parking spots, and thus cannot learn to ignore cluttering scene elements.


\section{Conclusions} \label{sec:conclusions}
In this work we have described CMRNet, a \ac{CNN} based approach for camera to LiDAR-Map registration, using the KITTI dataset for both learning and validation purposes.
The performances of the proposed approach allow multiple specialized CMRNet to be stacked as to improve the final camera localization, yet preserving realtime requirements.
The results have shown that our proposal is able to localize the camera with a median of less than $0.27$m and $1.07^\circ$.
Preliminary and not reported experiments on other datasets suggests there is room for improvement and the reason seems to be due to the limited vertical field-of-view available for the point clouds.
Since our method does not learn the map but learn how to perform the registration, it is suitable for being used with large-scale HD-Maps.

\section{Future Works} \label{sec:future-works}
Even though our approach does not embed any information of specific maps, a dependency on the intrinsic camera calibration parameters still holds.
As part of the future works we plan to increase the generalization capabilities so to not directly depend from a specific camera calibration.
Finally, since the error distributions reveal a similarity with respect to Gaussian distributions, we expect to be able to benefit from standard filtering techniques aimed to probabilistically tackle the uncertainties over time.

\section*{Acknowledgments}\label{sec:ack}
The authors would like to thank Tim Caselitz for his contribution related to the ground truth SLAM-based trajectories for the KITTI sequences and Pietro Colombo for the help in the editing of the associated video.

\bibliographystyle{IEEEtran}
\bibliography{IEEEabrv,dnn.loc.singleimage}

\end{document}